# HOW CAN AI FIND MY MODEL? A MODEL-FINDING EXPERIMENTAL STUDY CONSIDERING DATA FORMATS, EMBEDDINGS, AND RETRIEVAL STRATEGIES

Jhon G. Botello[1,2], Jose J. Padilla[2], Erika Frydenlund[2], Krzysztof Rechowicz[3], and Eric Weisel[4]

[1]Dept. of Computer Science, Old Dominion University, Norfolk, VA, USA
[2]Virginia Modeling, Analysis, and Simulation Center, Old Dominion University, Norfolk, VA, USA
[3]Virginia Digital Maritime Center, Old Dominion University, Norfolk, VA, USA
[4]Office of Enterprise Research and Innovation, Old Dominion University, Norfolk, VA, USA

## ABSTRACT

Discovering simulation models for reuse remains a fundamental challenge in Modeling and Simulation (M&S). When many models coexist, identifying those that align with a given modeling intent remains difficult. Recent advances in Artificial Intelligence (AI), particularly retrieval-based approaches, offer a promising pathway to operate at this semantic layer. In this paper, we present an experimental study investigating the impact of data representation, transformer-based embedding models, and retrieval strategies on the discovery of simulation models using natural language queries. We evaluated performance across multiple query types using standard information retrieval metrics, including recall@5 and nDCG@5. Results show that data representation matters, open-source embedding models can achieve high performance, and reranking methods are important, especially as query complexity increases. This work provides a baseline for AI-driven model discovery and discusses its role in advancing toward AI-driven composability and interoperability.

## 1 INTRODUCTION

Searching for and selecting models for reuse or composition remains complex, expensive, and labor-intensive. Modelers must navigate disparate repositories and documentation to match model capabilities with modeling questions, delaying readiness and increasing costs. The reuse of models inherently depends on the ability to discover relevant artifacts, which requires identifying and selecting models based on characteristics and intended use (Zeigler 2022). Conceptual modeling supports this process by enabling the mapping between user objectives and model components (Rheinsmith et al. 2010). In parallel, model discovery in M&S has traditionally relied on lexical retrieval (López and Cuadrado 2022) and ontology-based approaches (Mejdal et al. 2021). While Lexical methods limit semantic mapping, ontology-based approaches require manual knowledge engineering, limiting scalability and generalizability.

Discovering suitable simulation models also remains challenging when the objective extends to combining models and/or interoperable simulations. Tolk (2024) argues that, for successful simulation interoperability, the exchange of information at the technical level is not sufficient; received information must also be correctly understood and must trigger the intended behavior under the appropriate assumptions and constraints. As such, searching for and selecting models with a high likelihood of semantic alignment in the context of a modeling question should facilitate their identification, even if composability or interoperability evaluation is done manually. However, automating this search-and-selection process based on a modeling question is likely an undecidable problem. Weisel (2004) argues that determining whether elements of one model can be meaningfully mapped to another may exhibit undecidable properties as compatibility depends not only on structural correspondence but also on assumptions, abstractions, and behaviors. Accordingly, if we consider a modeling question as a model, identifying a simulation artifact to which some of the elements of the modeling question can be mapped may not be guaranteed to be fully automated. A modeling question, in this work, adapts Tolk et al. (2013) and Diallo et al. (2014) definitions as a collection of sentences that can be satisfied within a reference model.

Recent advances in AI, however, particularly retrieval-based approaches (e.g., Karpukhin et al. 2020; Nogueira and Cho 2020), offer a potential pathway to facilitate the semantic identification of models. We posit that while AI will not solve the undecidability problem, it provides an inductive approach for ingesting data and evaluating different forms of retrieval approaches, offering practical advantages, including reduced manual effort in model search and discovery. In this work, we frame simulation model discovery as an information retrieval (IR) problem and explore how pre-trained, transformer-based embeddings can provide a semantic layer to facilitate modelers' exploration of candidate simulations for a modeling question expressed as a natural-language query. To this end, we conducted an experimental study to evaluate data representations, embedding models, and ranking approaches. We provide a foundational study focused on the capabilities and limitations of current AI-based retrieval in isolation in M&S. This allows us to establish baseline performance across different configurations and representations before introducing additional semantic structures. We demonstrate our approach in a system dynamics (SD) setting and discuss its extension to other simulation paradigms. Our results highlight how AI can support modelers and contribute to the discussion on advancing toward AI-driven composability and interoperability.

## 2 BACKGROUND

Model discovery is the process of finding desired models or model components for reuse and is the first step toward reusing simulation artifacts (Hussain et al. 2022). Model discovery has long been recognized as an important challenge in M&S (Hussain et al. 2022; Kyurkchiev et al. 2026; Szabo and Teo 2011). Key challenges include the lack of appropriate model representations, the difficulty of ranking partial matches to a user query, and the absence of meaningful measures of similarity across models (Szabo and Teo 2011).

Previous work in M&S has explored syntactic discovery approaches based on keyword search over textual descriptions, model attributes, and metadata (López and Cuadrado 2022; Stone et al. 2011). While these approaches improve access to models, they are limited to lexical matching and lack robust mechanisms to capture semantic relationships, which are necessary for effective model retrieval in practice. To address these limitations, semantic discovery approaches have been proposed using ontologies, in which models are matched based on entity relationships, reference functions, structural attributes, and data types represented as part of domain knowledge (Mejdal et al. 2021; Moradi et al. 2007). Although ontologies provide a structured and interpretable way to represent domain knowledge, they typically require significant manual effort to design, maintain, and extend. Their effectiveness depends on the availability of well-defined vocabularies and consistent annotations, which are costly to produce and difficult to scale. Furthermore, ontology-based approaches rely on explicitly defined relationships between concepts, which limits their ability to capture implicit similarities. For example, semantically related modeling concepts such as "resource allocation" and "capacity planning" may not be recognized as similar unless their relationship has been explicitly encoded. This dependence on predefined relationships may reduce flexibility when dealing with heterogeneous model representations, domains, conceptualization, and variations in how modeling intent is expressed.

AI, particularly embedding-based representations, has opened new ways to address model discovery. Kyurkchiev et al. (2026) investigated a semantic search method for system dynamics models using vector embeddings, showing improved retrieval performance over traditional keyword-based techniques. In their work, models represented as graphs are transformed into textual descriptions to enable embedding-based similarity and semantic matching between queries and models. However, it is still necessary to explore different data representations, embedding models, preprocessing strategies, and retrieval pipelines as model discovery performance depends on how semantic information is encoded and retrieved, which can vary significantly across different query types and levels of abstraction. Such work will provide baseline strategies for model discovery that can be compared against alternative semantic methods in future work, while also informing scalability strategies and potential extensions to heterogeneous simulation paradigms. In our work, we depart from approaches that transform models into purely textual descriptions and instead operate directly on their native XML-based structure (e.g., XMILE representations for system dynamics models). This allows us to preserve the original structural and semantic information encoded in the model without introducing transformation-induced loss or ambiguity. We investigate how different data formats—ranging from raw XML-based representations to transformations such as JSON-LD and JSON-LD enriched with descriptions from metadata—affect retrieval performance. We provide an experimental evaluation across multiple configurations, including the data formats mentioned above, combined with lexical approaches and dense retrieval methods, with variations in embedding models, chunking size, overlap, and ranking pipelines. These pipelines include baseline dense retrieval using cosine similarity, as well as enhanced approaches that incorporate reranking via a cross-encoder and a large language model (LLM). By isolating and comparing these factors, this study aims to provide a deeper understanding of how representation and retrieval design choices influence the effectiveness of semantic model discovery, particularly across different levels of complexity in modeling questions, while offering insights into scalability and extension to heterogeneous simulation contexts.

### 2.1 Information Retrieval Background

Information retrieval (IR) is a well-established field, primarily grounded in computer science and closely related to information and library science. It provides a formal framework for retrieving relevant documents in response to a user query. Under this view, relevance is determined by how well a candidate document aligns with the intent expressed in a query (Robertson and Zaragoza 2009). Traditional IR approaches rely on lexical matching techniques, where documents are represented as collections of terms and retrieval is based on keyword overlap. Some of the most relevant methods include TF-IDF (Ramos 2003), BM25 (Robertson and Zaragoza 2009), and query likelihood (Ponte and Croft 2017). While these methods are computationally efficient and provide a baseline, they are often limited in their ability to capture semantic relationships, particularly when queries and documents use different vocabulary to express semantically similar concepts. To address these limitations, dense retrieval (Karpukhin et al. 2020) emerged, leveraging embedding-based representations to encode both queries and documents into a shared vector space. In this space, similarity is computed based on distance metrics such as cosine similarity, enabling the retrieval of semantically related items even in the absence of exact term overlap.

Further retrieval architectures include a two-stage pipeline in which an initial retriever produces a candidate set and a second model reranks those candidates. Cross-encoder rerankers have been shown to improve ranking quality by jointly encoding the query and candidate text and directly scoring their relevance, rather than scoring them independently in separate embedding spaces (Nogueira and Cho 2020). More recently, LLMs have also been explored as rerankers. Rather than relying only on fixed similarity functions, LLM-based reranking can use instruction-following and broader language understanding to compare candidates in context. Sun et al. (2023) show that properly prompted LLMs can serve as competitive reranking agents on established IR benchmarks, suggesting that they may be useful when relevance judgments require richer semantic interpretation. Evaluation in IR commonly relies on retrieval metrics. Recall is commonly used to measure the proportion of relevant items successfully retrieved. For ranked outputs, gain-based metrics are especially useful because they account for the position of relevant items in the returned list. Järvelin and Kekäläinen (2002) Introduced the cumulated gain and normalized discounted cumulative gain (nDCG) to evaluate ranked retrieval under graded relevance, making these measures well-suited for settings in which some retrieved documents may be more relevant than others. Although IR is a well-known domain, its application in M&S is still limited.

### 2.2 System Dynamics as an Application Domain

We selected SD as the application domain for this study because models are available in XML-based formats. XML is important as it provides a broadly adopted structure that is understandable by both humans and computers. Further, XML use for data exchange provides the means to evaluate composability/interoperability for candidate models. There has been a push toward standardization through common interchange formats in the SD software tools. One widely adopted standard is XMILE (OASIS 2015), an XML-based format designed to represent system dynamics models in a platform-independent manner. XMILE encodes key model components, including variables, equations, simulation specifications, and structural relationships, enabling models to be shared and processed across tools. This common representation facilitates interoperability and reproducibility. As such, SD provides a structured foundation for computational analysis and for exploring model discovery at scale, while generating insights into how this work can be extended to other simulation paradigms.

## 3 METHOD

This study systematically evaluates the impact of data representation in combination with retrieval configurations for simulation model discovery. By data representation, we refer to how data is structured and encoded, and how these choices influence retrieval performance. By retrieval configurations, we mean the methods used to index, retrieve, and rank candidate models in response to a query. To this end, we designed a set of controlled experiments across two main paradigms: lexical retrieval and dense retrieval (Figure 1). Code and data are available on GitHub to ensure reproducibility.

### 3.1 Data Collection and Preparation

Before evaluating any configuration, we collected and prepared a corpus of simulation models. As mentioned above, this study focuses on SD artifacts in their XML-based representation (XMILE), which can be exported directly from modeling platforms. A key challenge in our study is the absence of a publicly available, curated repository of simulation models. Unlike traditional IR contexts, simulation models are distributed across multiple websites and often reside in unmaintained repositories. As a result, they may be outdated in format and frequently lack consistent or complete metadata, requiring manual identification, selection, and curation. When available, metadata typically includes a model name and a brief textual description of the modeled system. We compiled a dataset of 40 system dynamics models with basic metadata, obtained from educational platforms such as Shodor and Forio. The collected models were developed using Vensim; therefore, each model was updated to the last version of Vensim and exported in XMILE format. This file represents the first data representation explored for model discovery. We note that although our dataset is based on Vensim models, the proposed approach is not restricted to a specific SD modeling tool. Any SD model exported as XMILE can be incorporated into our retrieval experiment.

While the XML-based format provides a standardized, machine-readable representation, it is inherently verbose since it encodes both structural and syntactic details that may obscure higher-level semantics relevant to model discovery. This limitation motivated us to explore an alternative representation to reduce verbosity while preserving essential model elements. We developed a parsing and transformation algorithm based on a set of rule-based heuristics designed for XMILE files. The transformation process extracts key semantic elements, such as simulation specifications, variables, equations, and their relationships, and encodes them into a compact JSON-LD representation, which remains machine-readable while reducing syntactic noise. It constitutes our second data representation. Additionally, we enrich the JSON-LD representation with textual descriptions from available metadata, creating a third artifact that enables evaluation of the impact of semantic augmentation on retrieval performance.

### 3.1 Lexical Retrieval

Lexical retrieval was used as a baseline and includes traditional term-matching techniques such as TF-IDF, BM25, and query likelihood. As defined in (1), TF-IDF assigns weights to terms based on their frequency within a document and their inverse frequency across the collection, where $tf(t,d)$ denotes the frequency of the term $t$ in document $d$, $df(t)$ is the number of documents containing $t$, and $N$ is the total number of documents in the collection. In our case, a document corresponds to a simulation model in any specified data representation.

$$TF\text{-}IDF(t,d) = tf(t,d) \cdot \log\left(\frac{N}{df(t)}\right) \tag{1}$$

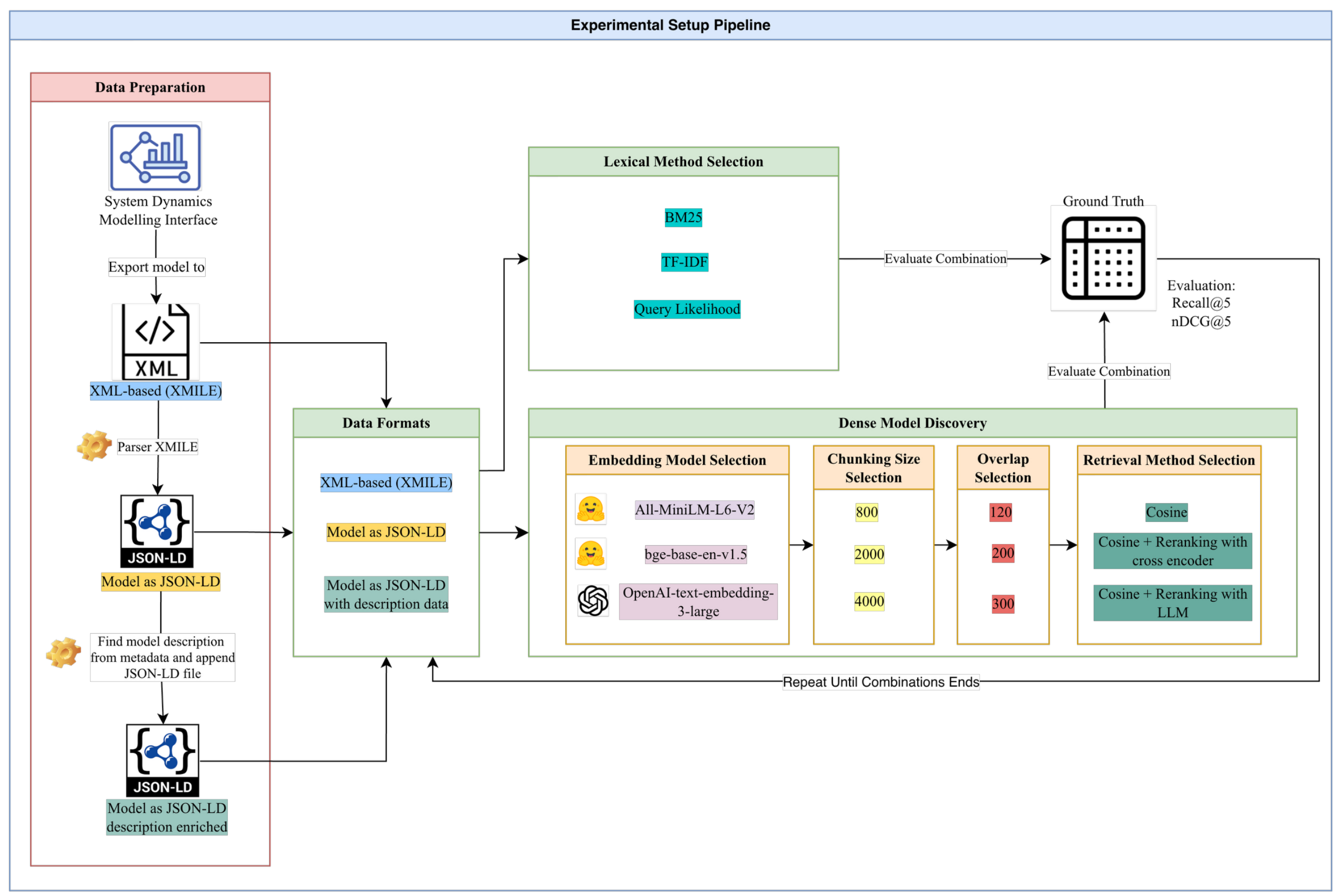

Figure 1. Experimental pipeline for simulation model discovery

BM25 (2) extends TF-IDF (1) by incorporating term-frequency saturation and document-length normalization. $f(q_i, D)$ represents the frequency of the query term $q_i$ in the document $D$, $|D|$ is the document length, and $avgdl$ is the average document length across the collection. The parameters $k_1$ and $b$ control term frequency saturation and length normalization, respectively.

$$\text{score}(D, Q) = \sum_{q_i \in Q} IDF(q_i) \cdot \frac{f(q_i, D)(k_1 + 1)}{f(q_i, D) + k_1 \left(1 - b + b \cdot \frac{|D|}{avgdl}\right)} \tag{2}$$

The query likelihood approach follows a probabilistic language method defined by $P(d \mid q) \propto P(q \mid d)$, where documents are ranked according to the probability of generating the query $q$ given the document $d$. In this approach, each document is treated as a probabilistic model over terms, and query terms are assumed to be generated independently.

We adopted an experimental setting with two factors: (1) data representation (XML, JSON-LD, and JSON-LD with narratives) and (2) lexical method (BM25, TF-IDF, and query likelihood), resulting in 9 possible configurations. Simulation models were indexed at the document level, as lexical retrieval operates on term-based representations and can efficiently handle full document content. All documents and queries were lowercased and tokenized using a regular-expression tokenizer. TF-IDF used a bag-of-words representation with dot-product similarity, BM25 was implemented via the rank-bm25 Python library (Brown 2022), and query likelihood used a unigram language model with Dirichlet smoothing (μ = 1000). For all configurations, the top-k documents were retrieved with k = 5.

### 3.2 Dense Retrieval

We evaluated dense retrieval under variations of embedding models, chunk size, chunk overlap, and ranking method. The last one included dense-only ranking and reranking approaches using a cross-encoder and an LLM. Unlike lexical retrieval, which relies on exact term matching, dense retrieval encodes both queries and documents into a shared continuous vector space. Due to the input length limitations of embedding models, documents were segmented into overlapping chunks prior to indexing. This decision allows scalability. As such, given a query $q$ and a document (chunk) $d$, relevance is computed using cosine similarity, as defined in (3), where $\mathbf{q}$ and $\mathbf{d}$ denote the embedding vectors of the query and document, respectively. In our implementation, all embeddings are normalized to unit length, making cosine similarity equivalent to the dot product:

$$\mathrm{sim}(q, d) = \frac{\mathbf{q} \cdot \mathbf{d}}{\| \mathbf{q} \| \| \mathbf{d} \|} \tag{3}$$

We applied a sliding window approach with chunk sizes $c \in \{800, 2000, 4000\}$ and overlaps $o \in \{120, 200, 300\}$, ensuring that long simulation models are fully represented while preserving local context. Each chunk is treated as an independent retrieval unit and encoded as a vector. We explored the performance of three embedding models, including *all-MiniLM-L6-v2* (Wang et al. 2020), *BAAI/bge-base-en-v1.5* (Xiao et al. 2024), and *text-embedding-3-large* (OpenAI 2024). All embeddings are normalized to unit vectors to ensure consistent similarity computation across models.

Chunk embeddings were indexed using a vector database (Chroma DB) with cosine distance, and for each query, the top-$k$ most similar chunks are retrieved, with $k_{\text{chunks}} = 20$. This number can be varied in future studies. Since retrieval operates at the chunk level in our study to allow large models to be represented and to support scalability, document-level rankings are obtained by aggregating chunk scores. We used a weighted sum formulation, as defined in (4), where $s_i$ denotes the similarity score of the $i$-th ranked chunk associated with the document $D$, and $\alpha = 0.7$ is a decay factor that prioritizes higher-ranked chunks:

$$\mathrm{score}(D) = \sum_{i=1}^{m} \alpha^{i-1} \cdot s_i \tag{4}$$

The final ranking returns the top-$k$ documents, with $k_{\text{docs}} = 5$, enabling direct comparison with the lexical retrieval setting. To further analyze ranking strategies, we evaluated three methods. The first configuration (dense-only) ranks documents directly based on aggregated similarity scores. The second approach (dense + reranking) refines the top-$N = 20$ retrieved chunks using the cross-encoder model *cross-encoder/ms-marco-MiniLM-L-6-v2* (Hugging Face 2025), which has already been trained on the MS MARCO passage ranking dataset (Bajaj et al. 2018). The third approach (dense + LLM reranking) considers the top-$N = 10$ chunks and applies a large language model (*gpt-4o-mini*) to assign relevance scores on a scale of $[0, 100]$ based on query–chunk alignment. Continuous scoring ranges like the one we use have been commonly employed in LLM-based evaluation settings to provide fine-grained judgments (Gu et al. 2026). A structured prompt is given to the LLM to perform the task. The prompt includes the four elements identified by Giray (2023) as essential for a good prompt: instructions, context, input data, and an output indicator. By combining all experimental factors (data representation, embedding models, chunking sizes, overlaps, and ranking modes), we evaluated 243 configurations.

### 3.3 Ground Truth and Evaluation Metrics

Given the absence of a publicly available benchmark dataset for simulation model discovery, one of our team members with a strong background in SD manually annotated 15 queries with graded relevance labels to reflect the relative importance of each simulation model. Only a subset of models was used for annotation, while the remaining models were treated as noise to better simulate a real-world retrieval scenario. To support a diverse and representative evaluation, queries were designed across three categories with increasing levels of complexity: direct, abstract, and ambiguous. Direct queries explicitly mention specific concepts present in relevant models. Abstract queries describe higher-level concepts, including synonymous or semantically related terms. Ambiguous queries are intentionally underspecified, targeting more complex structures or dynamics within models and requiring broader semantic interpretation. We posit that a query representing a natural-language modeling question can be viewed as an abstraction of modeling requirements, including concepts, entities, relationships, and behaviors. For example, a direct query like *"Which simulation models represent ecological population dynamics using explicit stock variables and flows such as births, deaths, growth, or competition to analyze population behavior under resource or environmental constraints?"* encodes multiple elements: entities (e.g., population, resources, environment), concepts (ecological population dynamics), relationships and behavior (interactions between population and constraints), and properties (births, deaths, growth, competition, represented as stocks and flows). We evaluated each configuration using the annotated ground Truth for both lexical and dense retrieval, implementing Recall@5 and nDCG@5. Recall@5 measures how many of the relevant models appear in the top 5 results. nDCG@5 considers both relevance and ranking, placing greater weight on relevant models that appear earlier in the results.

## 4 RESULTS

To enable a structured comparison, we report the mean performance and the top three configurations for each data representation (JSON-LD with narratives, JSON-LD, and XMILE) within each query type (direct, abstract and ambiguous) for both Lexical (Table 1) and dense (Table 2) retrieval.

Table 1. Top three lexical retrieval configurations across query types and data representations

| Query Type | | Mean Recall@5 | Mean nDCG@5 | Top Three Experiments performance | Recall@5 | nDCG@5 |
|---|---|---|---|---|---|---|
| Direct | JSON-LD-Narrative | 0.78 | 0.67 | json_narratives__tfidf | 0.80 | 0.68 |
| | | | | json_narratives__bm25 | 0.73 | 0.67 |
| | | | | json_narratives__query_likelihood | 0.80 | 0.66 |
| | JSON-LD | 0.56 | 0.37 | json__bm25 | 0.50 | 0.33 |
| | | | | json__tfidf | 0.50 | 0.34 |
| | | | | json__query_likelihood | 0.67 | 0.43 |
| | XMILE | 0.42 | 0.29 | xml__bm25 | 0.34 | 0.25 |
| | | | | xml__tfidf | 0.40 | 0.29 |
| | | | | xml__query_likelihood | 0.53 | 0.33 |
| Abstract | JSON-LD-Narrative | 0.27 | 0.24 | json_narratives__bm25 | 0.28 | 0.31 |
| | | | | json_narratives__tfidf | 0.28 | 0.23 |
| | | | | json_narratives__query_likelihood | 0.23 | 0.19 |
| | JSON-LD | 0.05 | 0.04 | json__query_likelihood | 0.05 | 0.06 |
| | | | | json__bm25 | 0.05 | 0.03 |
| | | | | json__tfidf | 0.05 | 0.03 |
| | XMILE | 0.14 | 0.09 | xml__bm25 | 0.18 | 0.12 |
| | | | | xml__tfidf | 0.18 | 0.11 |
| | | | | xml__query_likelihood | 0.05 | 0.03 |
| Ambiguous | JSON-LD-Narrative | 0.10 | 0.08 | json_narratives__tfidf | 0.10 | 0.12 |
| | | | | json_narratives__query_likelihood | 0.10 | 0.09 |
| | | | | json_narratives__bm25 | 0.10 | 0.03 |
| | JSON-LD | 0.03 | 0.01 | json__bm25 | 0.05 | 0.01 |
| | | | | json__tfidf | 0.00 | 0.00 |
| | | | | json__query_likelihood | 0.05 | 0.01 |
| | XMILE | 0.06 | 0.03 | xml__bm25 | 0.12 | 0.06 |
| | | | | xml__tfidf | 0.07 | 0.03 |
| | | | | xml__query_likelihood | 0.00 | 0.00 |

For lexical retrieval, the best performance was on direct queries, using JSON-LD, particularly enriched with metadata (JSON-LD narratives) and using TF-IDF, while XMILE performs worse. This suggests that JSON-based representations better align with the keyword-driven nature of direct queries. Metadata enrichment likely improves term matching by increasing lexical overlap, while even without narratives, JSON might reduce structural noise and enhance term salience compared to XML, leading to improved retrieval effectiveness. For abstract and ambiguous queries, overall performance drops considerably across all configurations, confirming the limitations of lexical retrieval in capturing non-explicit information needs. Notably, for this type of query, while JSON_narratives achieves higher performance, XML-based representations sometimes outperform JSON alone. We argue that this behavior might be due to XML tags. For instance, if a query mentions retrieving “stock” variables, the JSON-LD representation may include a single “stock” field listing all variables, whereas the term “stock” may appear multiple times in the XML-based file. Overall, while baseline, the sharp decline in performance from direct to abstract and ambiguous queries confirms that purely lexical approaches are insufficient for simulation model discovery.

Results from dense retrieval show clear performance improvements across all query types compared to lexical approaches. The results suggest that it is possible to achieve high performance in model discovery using a local embedding model when queries are specific, such as direct queries. Yet reranking would still be necessary. As query complexity increases, enriching models with metadata becomes important to improve performance, although challenges remain, as the drop in recall suggests. This behavior can be explained by the fact that, without metadata, embedding-based representations rely primarily on structural information, which may fail to capture higher-level semantics or implicit modeling intent, as this would require inferences about model behavior. Additionally, in the presence of noisy models, abstract or ambiguous queries may match irrelevant models, as there is no explicit representation of variable or concept behavior, and a similarity match does not imply inference.

To illustrate the differences and challenges in retrieval, we provide an example. Figure 2 shows the relevant models for the query introduced in Section 3.3, where (a) is more relevant than (b), and (b) is more relevant than (c).

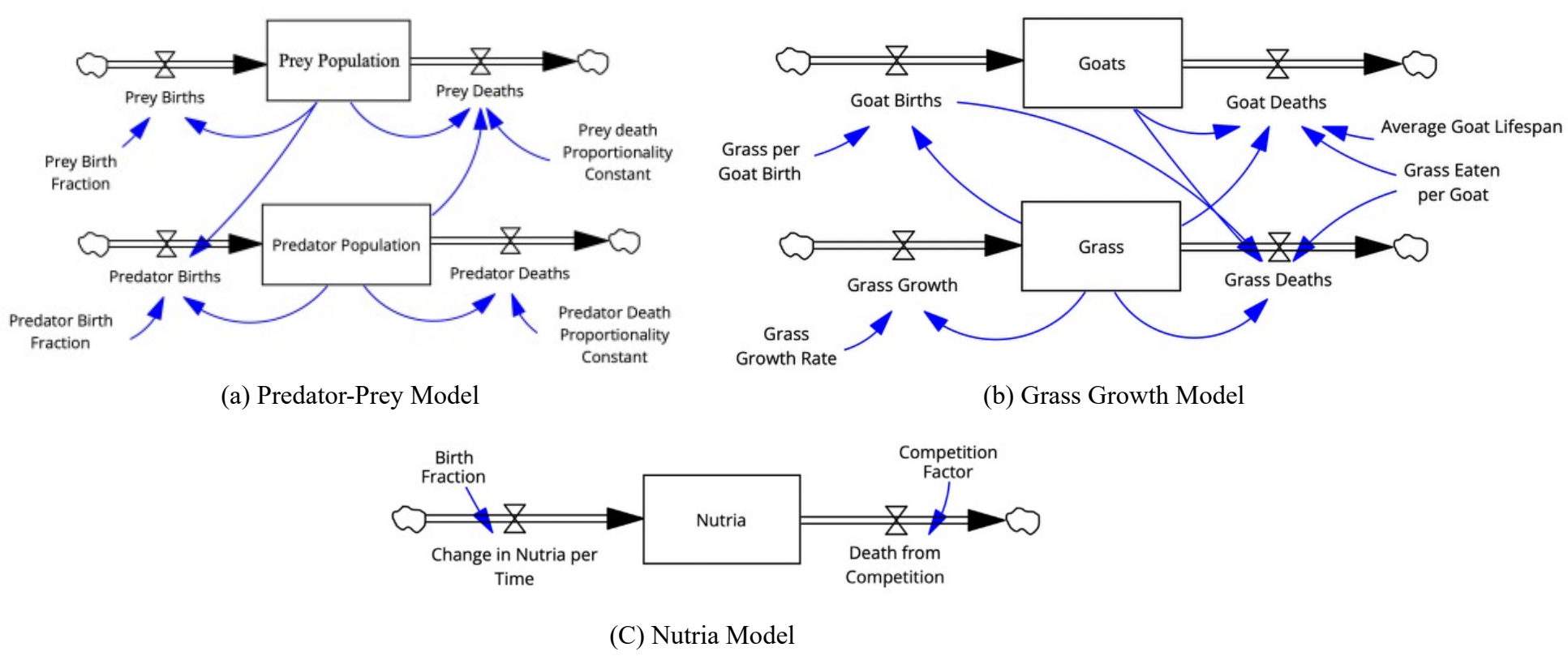

Figure 2. Relevant models for query introduced in section 3.3.

When using the direct query, configurations based on JSON-LD may be able to retrieve the relevant models, as key terms such as birth, death, and population are explicitly present. However, ranking improves when narrative descriptions are included, as they might help to better match the intent of the modeling question. As the query becomes more abstract or ambiguous—for example, when simplified to refer only to population dynamics—the retrieval task would require understanding the underlying interactions of the variables and system, rather than relying on only semantically similar terms. This makes it more difficult to distinguish between highly relevant and partially relevant models.

The above is supported by the retrieval challenges observed when metadata is absent, where JSON and XMILE show similarly challenging performance, with XMILE slightly outperforming in some cases. We contend that the advantage of transforming XML-based representations into JSON lies in reducing token consumption. Dense retrieval is computationally expensive, and larger representations increase both processing time and embedding and reranking cost, particularly when using API-based models. For instance, in our study, the full dense experimental pipeline evaluated 243 configurations in 2.45 hours, with an average of 36.26 seconds per experiment and 2.42 seconds per query in a machine with 64 GB RAM and no GPU installed. Regarding the use of OpenAI's API, embedding generation cost $0.55 for 4.25M tokens, while reranking cost $0.43 across 4,050 API calls, totaling approximately $0.99. Although this cost is modest, it reflects a single indexing and evaluation pass for experimentation. In practice, costs (both computational and monetary) can scale significantly with larger repositories, frequent queries, and the addition of generation steps in RAG pipelines. The above opens the door to exploring how open source LLMs perform for reranking in this M&S task, potentially leading toward finetuning.

Table 2. Top three dense retrieval configurations across query types and data representations.

| Query Type | | Mean Recall @5 | Mean nDCG @5 | Top Three Experiments performance | Recall @5 | nDCG @5 |
|---|---|---|---|---|---|---|
| Direct | JSON-LD-Narrative | 0.71 | 0.59 | json-narratives_cs4000_ov300_bge_LLMRerank | 0.93 | 0.89 |
| | | | | json-narratives_cs4000_ov200_bge_ LLMRerank | 0.93 | 0.82 |
| | | | | json-narratives_cs4000_ov120_bge__ LLMRerank | 0.93 | 0.81 |
| | JSON-LD | 0.59 | 0.44 | json-cs800_ov200_openai_ LLMRerank | 0.87 | 0.64 |
| | | | | json_cs800_ov300_openai_ LLMRerank | 0.80 | 0.63 |
| | | | | json_cs2000_ov120_openai_ LLMRerank | 0.93 | 0.63 |
| | XMILE | 0.54 | 0.43 | xml_cs4000_ov300_openai__ LLMRerank | 0.60 | 0.62 |
| | | | | xml_cs800_ov200_openai_LLMRerank | 0.73 | 0.61 |
| | | | | xml_cs2000_ov120_openai_LLMRerank | 0.73 | 0.62 |
| Abstract | JSON-LD-Narrative | 0.33 | 0.27 | json-narratives_cs4000_ov120_openai_LLMRerank | 0.50 | 0.48 |
| | | | | json-narratives_cs4000_ov200_openai_LLMRerank | 0.50 | 0.48 |
| | | | | json-narratives_cs4000_ov300_openai_LLMRerank | 0.50 | 0.48 |
| | JSON-LD | 0.29 | 0.22 | json_cs800_ov300_bge_CrossencoderRerank | 0.35 | 0.49 |
| | | | | json_cs4000_ov200_bge_CrossencoderRerank | 0.35 | 0.38 |
| | | | | json_cs2000_ov120_bge_CrossencoderRerank | 0.33 | 0.37 |
| | XMILE | 0.23 | 0.17 | xml_cs2000_ov200_minilm_CrossencoderRerank | 0.36 | 0.41 |
| | | | | xml_cs4000_ov200_bge_CrossencoderRerank | 0.35 | 0.40 |
| | | | | xml_cs800_ov120_bge_ LLMRerank | 0.30 | 0.39 |
| Ambiguous | JSON-LD-Narrative | 0.33 | 0.36 | json-narratives_cs2000_ov120_minilm_NoRerank | 0.37 | 0.69 |
| | | | | json-narratives_cs800_ov300_minilm_NoRerank | 0.42 | 0.65 |
| | | | | json-narratives_cs800_ov200_minilm_NoRerank | 0.32 | 0.63 |
| | JSON-LD | 0.17 | 0.20 | json_cs2000_ov120_minilm_LLMRerank | 0.25 | 0.44 |
| | | | | json_cs2000_ov200_minilm_LLMRerank | 0.25 | 0.44 |
| | | | | json_cs4000_ov120_minilm_LLMRerank | 0.25 | 0.43 |
| | XMILE | 0.15 | 0.16 | xml_cs4000_ov300_bge_CrossencoderRerank | 0.31 | 0.47 |
| | | | | xml_cs4000_ov120_bge_LLMRerank | 0.31 | 0.46 |
| | | | | xml_cs4000_ov200_bge_ CrossencoderRerank | 0.26 | 0.44 |

Overall, results from dense retrieval show that it can accelerate search and model discovery, especially when combined with LLM-based reranking. However, performance still depends on having well-defined queries and enriching models with metadata. These findings suggest the need for structured templates for model indexing. For example, representing models in JSON-LD while transforming key elements—such as concepts, variables, relationships, and parameters—into narrative templates may further improve retrieval performance. This could be explored using LLMs or rule-based methods that extract components from the XML-based structure and populate the template.

# 5 DISCUSSION AND LIMITATIONS

Results highlight the importance of incorporating descriptions into models and clarity in questions. This means that simulation repositories should prioritize rich model descriptions and metadata. Narratives provide context to match modeling questions to models, seemingly bounding the undecidability challenge by providing a mechanism to reduce the search space and guide modelers toward promising candidates, especially when combined with the use of LLMs as rerankers, as they enable inference grounded in broader contextual knowledge learned during training. For modelers, AI-assisted retrieval can reduce the effort required to identify relevant models, accelerate reuse workflows, and provide a smaller set of semantically aligned candidates for manual evaluation. It also may support more informed decision-making during model reuse, and further composability or interoperability analysis. However, accelerating the evaluation of composability after model discovery remains an open area. Although it can also be framed as an undecidable problem, exploring the advantages of LLMs for this task—or rule-based methods grounded in embeddings—represents a promising direction for helping to bound the problem.

Looking forward, several avenues for future research emerge from this study. First, future work can explore hybrid retrieval techniques that combine dense, lexical, and graph-based methods such as graph RAG. Such approaches may further improve retrieval. Second, our work can be extended to other modeling paradigms. For instance, the parsing approach proposed in this study could be expanded to identify semantic characteristics in agent-based models developed in Netlogo, such as agents properties, interaction rules, state variables, and environment structures by leveraging formats such as .nlogox, an XML-based format released with the 7th version of NetLogo in September 2025 (NetLogo 2025). Exploring this avenue may enable AI-driven cross-paradigm model discovery. The above also creates opportunities for comparative and integrative model exploration, where users can identify functionally similar models despite differences in formalism. Third, from a broader perspective, our work also contributes to emerging efforts on rapid simulation development (e.g., Botello et al. 2025; Martínez et al. 2024), where the goal is to reduce the time and effort required to construct simulation models from high-level descriptions. In these settings, RAG-based architectures can provide initial models to build upon, while the experimental findings reported here can inform the design of retrieval-augmented systems by offering guidance on data representations, embedding models, and ranking strategies for simulation model discovery.

We highlight several limitations of our work. First, our evaluation was based on a relatively small query set (15 queries) over 40 simulation models, which may limit generalizability, particularly for abstract and ambiguous queries that exhibit high variability. However, the observed trends—namely the relative improvements of dense over lexical retrieval and the impact of metadata enrichment and reranking—are consistent with well-established findings in information retrieval, suggesting that these patterns are likely to hold directionally at larger scales. Additionally, the challenges of collecting, curating, and annotating simulation models make this study an important first step toward the development of large-scale benchmark datasets for model discovery. Such benchmarks will be essential for evaluating more advanced retrieval methods and scaling experiments to larger scenarios. Second, our ground truth was generated by a single expert annotator, introducing potential subjectivity. Future work should involve multiple annotators and inter-annotator agreement measures to strengthen reliability. Finally, dense retrieval introduces notable computational costs. Therefore, leveraging more powerful computational resources should be considered to reduce processing time in future experiments.

## 6 CONCLUSION

This paper presented an experimental study on simulation model discovery, analyzing how data representations, embedding models, and retrieval strategies impact the ability to identify relevant models from natural language queries. Results show that representation, metadata enrichment and reranking play a key role in retrieval effectiveness, especially as query complexity increases. However, challenges remain in capturing abstract and ambiguous modeling intent, highlighting the inherent complexity of the problem. Overall, this work establishes a baseline for AI-driven model discovery and provides insights into scalable approaches and future strategies that support model reuse, rapid simulation development, composability, and interoperability in M&S.

## ACKNOWLEDGMENTS

We thank Dr. Andreas Tolk for his insights on the topics discussed in this paper.

## REFERENCES

Bajaj, P., Campos, D., Craswell, N., Deng, L., Gao, J., Liu, X., Majumder, R., McNamara, A., Mitra, B., Nguyen, T., Rosenberg, M., Song, X., Stoica, A., Tiwary, S., & Wang, T. (2018). MS MARCO: A Human Generated MAchine Reading COmprehension Dataset (arXiv:1611.09268). arXiv. https://doi.org/10.48550/arXiv.1611.09268

Botello, J. G., Llinas, B., Padilla, J. J., & Frydenlund, E. (2025). Toward Automating System Dynamics Modeling: Evaluating Llms in the Transition From Narratives to Formal Structures. 2025 Winter Simulation Conference (WSC), 2380–2391. https://doi.org/10.1109/WSC68292.2025.11338856

Brown. (2022). Rank-BM25: A two line search engine. https://pypi.org/project/rank-bm25/

Diallo, S. Y., Padilla, J. J., Gore, R., Herencia-zapana, H., & Tolk, A. (2014). Toward a formalism of modeling and simulation using model theory. Complexity, 19(3), 56–63. https://doi.org/10.1002/cplx.21478

Giray, L. (2023). Prompt Engineering with ChatGPT: A Guide for Academic Writers. Annals of Biomedical Engineering, 51(12), 2629–2633. https://doi.org/10.1007/s10439-023-03272-4

Gu, J., Jiang, X., Shi, Z., Tan, H., Zhai, X., Xu, C., Li, W., Shen, Y., Ma, S., Liu, H., Wang, S., Zhang, K., Lin, Z., Zhang, B., Ni, L., Gao, W., Wang, Y., & Guo, J. (2026). A survey on LLM-as-a-judge. The Innovation, 101253. https://doi.org/10.1016/j.xinn.2025.101253

Hugging Face. (2025). Cross-encoder/ms-marco-MiniLM-L6-v2. https://huggingface.co/cross-encoder/ms-marco-MiniLM-L6-v2

Hussain, M., Masoudi, N., Mocko, G., & Paredis, C. (2022). Approaches for Simulation Model Reuse in Systems Design—A Review. SAE International Journal of Advances and Current Practices in Mobility, 04(5), 1457–1471. https://doi.org/10.4271/2022-01-0355

Järvelin, K., & Kekäläinen, J. (2002). Cumulated gain-based evaluation of IR techniques. ACM Transactions on Information Systems, 20(4), 422–446. https://doi.org/10.1145/582415.582418

Karpukhin, V., Oguz, B., Min, S., Lewis, P., Wu, L., Edunov, S., Chen, D., & Yih, W. (2020). Dense Passage Retrieval for Open-Domain Question Answering. Proceedings of the 2020 Conference on Empirical Methods in Natural Language Processing (EMNLP), 6769–6781. https://doi.org/10.18653/v1/2020.emnlp-main.550

Kyurkchiev, P., Iliev, A., & Kyurkchiev, N. (2026). Semantic Search for System Dynamics Models Using Vector Embeddings in a Cloud Microservices Environment. Future Internet, 18(2), 86. https://doi.org/10.3390/fi18020086

López, J. A. H., & Cuadrado, J. S. (2022). An efficient and scalable search engine for models. Software and Systems Modeling, 21(5), 1715–1737. https://doi.org/10.1007/s10270-021-00960-4

Martínez, J., Llinas, B., Botello, J. G., Padilla, J. J., & Frydenlund, E. (2024). Enhancing GPT-3.5's Proficiency in Netlogo Through Few-Shot Prompting and Retrieval-Augmented Generation. 2024 Winter Simulation Conference (WSC), 666–677. https://doi.org/10.1109/WSC63780.2024.10838967

Mejdal, S., Penas, O., Barbedienne, R., Plateaux, R., Bisquay, M., & Brunet, J. (2021). Ontology-Based Search Engine For Simulation Models From Their Related System Function. INCOSE International Symposium, 31(1), 1145–1159. https://doi.org/10.1002/j.2334-5837.2021.00892.x

Moradi, F., Ayani, R., & Tan, G. (2007). A Rule-based Approach to Syntactic and Semantic Composition of BOMs. 11th IEEE International Symposium on Distributed Simulation and Real-Time Applications (DS-RT'07), 145–155. https://doi.org/10.1109/DS-RT.2007.32

Netlogo. (2025). Netlogo [Computer software]. https://docs.netlogo.org/7.0.0/versions.html#version-700-september-2025

Nogueira, R., & Cho, K. (2020). Passage Re-ranking with BERT (arXiv:1901.04085). arXiv. https://doi.org/10.48550/arXiv.1901.04085

OASIS. (2015). XML Interchange Language for System Dynamics (XMILE) Version 1.0 [Computer software]. https://docs.oasis-open.org/xmile/xmile/v1.0/xmile-v1.0.html
OpenAI. (2024). Vector embeddings: Text-embedding-3-large [Computer software]. https://platform.openai.com/docs/guides/embeddings
Ponte, J. M., & Croft, B. (2017). A language modeling approach to information retrieval. In ACM SIGIR Forum, 51. https://dl.acm.org/doi/pdf/10.1145/3130348.3130368
Ramos, J. (2003). Using TF-IDF to determine word relevance in document queries. Proceedings of the First Instructional Conference on Machine Learning, 1, 242, 29–48.
Rheinsmith, R., Wallace, J., Bizub, W., Ceranowicz, A., Cutts, D., Powell, E. T., Gustavson, P., Lutz, R., & McCloud, T. (2010). Joint Composable Object Model and LVC Methodology. Selected Papers Presented at MODSIM World 2009 Conference and Expo. https://ntrs.nasa.gov/citations/20100012881
Robertson, S., & Zaragoza, H. (2009). The Probabilistic Relevance Framework: BM25 and Beyond. Foundations and Trends® in Information Retrieval, 4(1–2), 1–174. https://doi.org/10.1561/1500000019
Stone, G. F. I., Greenberg, B., Daehler-Wilking, R., & Hunt, S. (2011). The Modeling and Simulation Catalog for Discovery, Knowledge and Reuse. Selected Papers and Presentations Presented at MODSIM World 2010 Conference and Expo. https://ntrs.nasa.gov/citations/20110012125
Sun, W., Yan, L., Ma, X., Wang, S., Ren, P., Chen, Z., Yin, D., & Ren, Z. (2023). Is ChatGPT Good at Search? Investigating Large Language Models as Re-Ranking Agents. Proceedings of the 2023 Conference on Empirical Methods in Natural Language Processing, 14918–14937. https://doi.org/10.18653/v1/2023.emnlp-main.923
Szabo, C., & Teo, Y. M. (2011). An approach to semantic-based model discovery and selection. Proceedings of the 2011 Winter Simulation Conference (WSC), 3054–3066. https://doi.org/10.1109/WSC.2011.6148006
Tolk, A. (2024). Conceptual alignment for simulation interoperability: Lessons learned from 30 years of interoperability research. SIMULATION, 100(7), 709–726. https://doi.org/10.1177/00375497231216471
Tolk, A., Diallo, S. Y., Padilla, J. J., & Herencia-Zapana, H. (2013). Reference modelling in support of M&S—foundations and applications. Journal of Simulation, 7(2), 69–82. https://doi.org/10.1057/jos.2013.3
Wang, W., Wei, F., Dong, L., Bao, H., Yang, N., & Zhou, M. (2020). MiniLM: Deep Self-Attention Distillation for Task-Agnostic Compression of Pre-Trained Transformers (Version 2). arXiv. https://doi.org/10.48550/ARXIV.2002.10957
Weisel, E. W. (2004). Models, Composability, and Validity [Old Dominion University Libraries]. https://doi.org/10.25777/43PV-GS24
Xiao, S., Liu, Z., Zhang, P., Muennighoff, N., Lian, D., & Nie, J.-Y. (2024). C-Pack: Packed Resources For General Chinese Embeddings (arXiv:2309.07597). arXiv. https://doi.org/10.48550/arXiv.2309.07597
Zeigler, B. P. (2022). A methodology to characterise simulation models for discovery and composition: A system theory-based approach to model curation for integration and reuse. International Journal of Simulation and Process Modelling, 19(1/2), 3. https://doi.org/10.1504/IJSPM.2022.130277

## AUTHOR BIOGRAPHIES

**JHON G. BOTELLO, M.Sc.** is a Ph.D candidate in Computer Science at Old Dominion University and a Graduate Research Assistant at VMASC. His research focuses on applying Machine Learning (ML) and AI in the fields of M&S and Web Science. His email address is jbote001@odu.edu, and his web page is https://jgbotello.github.io/.

**JOSE J. PADILLA, Ph.D.** is a Research Associate Professor at VMASC. His primary research focuses on advancing computational modeling methods toward increasing modeling accessibility across ages and disciplines. His research generates insight into topics ranging from forced migration to community resilience. His email address is jpadilla@odu.edu.

**ERIKA F. FRYDENLUND, Ph.D.** is a Research Associate Professor at VMASC at Old Dominion University. Her research employs computational modeling and ethnographic methods to analyze complex human behaviors and societal dynamics, with particular attention to population movements and community responses to change. Her email address is efrydenl@odu.edu.

**KRZYSZTOF J. RECHOWICZ, Ph.D.** is a Research Associate Professor at the Virginia Digital Maritime Center at Old Dominion University. His research applies computational modeling, simulation, and artificial intelligence to connect disparate systems and amplify human and technological capability in maritime and defense contexts. His email address is krechowi@odu.edu.

**ERIC WEISEL, Ph.D.** is Senior Associate Vice President in the Office of Enterprise Research and Innovation at ODU. His primary research centers on model composability and simulation interoperability. His email adress is eweisel@odu.edu.